\documentclass[conference]{IEEEtran}
\IEEEoverridecommandlockouts
\usepackage{amsmath,amssymb,amsfonts}
\usepackage{graphicx}
\usepackage{textcomp}
\usepackage{xcolor}
\usepackage{biblatex}
\usepackage{textcomp}
\usepackage{adjustbox}
\usepackage{multirow}
\usepackage{bbm}
\usepackage{array}
\usepackage[utf8]{inputenc}
\usepackage{algpseudocode,algorithm,algorithmicx}
\usepackage{comment}
\usepackage{mathtools}

\def\BibTeX{{\rm B\kern-.05em{\sc i\kern-.025em b}\kern-.08em
    T\kern-.1667em\lower.7ex\hbox{E}\kern-.125emX}}

\makeatletter
\newcommand{\vast}{\bBigg@{4}}
\newcommand{\Vast}{\bBigg@{5}}
\makeatother

\usepackage{fancyhdr}
\begin{document}

\title{A Robust Negative Learning Approach to Partial Domain Adaptation Using Source Prototypes}

\author{\IEEEauthorblockN{Sandipan Choudhuri, Suli Adeniye, Arunabha Sen}
\text{Arizona State University}\\
\{s.choudhuri, sadeniye, asen\}@asu.edu}

\maketitle

\begin{abstract}
This work proposes a robust Partial Domain Adaptation (PDA) framework that mitigates the negative transfer problem by incorporating a robust target-supervision strategy. It leverages ensemble learning and includes diverse, complementary label feedback, alleviating the effect of incorrect feedback and promoting pseudo-label refinement. Rather than relying exclusively on first-order moments for distribution alignment, our approach offers explicit objectives to optimize intra-class compactness and inter-class separation with the inferred source prototypes and highly-confident target samples in a domain-invariant fashion. Notably, we ensure source data privacy by eliminating the need to access the source data during the adaptation phase through a priori inference of source prototypes. We conducted a series of comprehensive experiments, including an ablation analysis, covering a range of partial domain adaptation tasks. Comprehensive evaluations on benchmark datasets corroborate our framework's enhanced robustness and generalization, demonstrating its superiority over existing state-of-the-art PDA approaches.
\end{abstract}


\section{Introduction}

\noindent
Supervised learning models, widely utilized for their remarkable performance in various applications \cite{choudhuri2018object, ganin2016domain,liu2021review}, depend significantly on access to extensive, annotated datasets. Their procurement, often costly and challenging, limits models' generalizability \cite{liu2021review}. Furthermore, a crucial assumption underpinning the effectiveness of supervised learning is that training and testing data originate from the same distribution. This expectation often proves unrealistic, causing models to fail in adequately generalizing across related but different domains due to the extensively studied domain shift problem \cite{torralba2011unbiased}. Unsupervised Domain Adaptation (UDA) \cite{li2020deep,ganin2016domain} offers a promising solution in this scenario. It circumvents the need for abundant labeled data by utilizing source domain knowledge to improve model performance on unlabeled target domain samples. However, standard UDA methods assume identical label spaces between domains, which often falls short in real-world scenarios.

Partial Domain Adaptation (PDA) offers a more pragmatic approach, allowing the source label space to subsume the target label space \cite{cao2018partial}. Nevertheless, PDA comes with its own challenges, primarily arising from the outlier classes in the source (labeled) domain. Aligning entire distributions between domains often results in ``negative transfer" (described in sec. \ref{prob_set}), where the classifier performance is adversely affected on target (unlabeled domain) data due to the unwarranted information from source \cite{cao2018partial,cao2019learning}. To counter this, proposed PDA strategies focus on filtering irrelevant source data by re-weighting sample predictions, aggregating category-level predictions, or by using averaged features as source prototypes coupled with adaptive thresholding  \cite{cao2018partial,zhang2018importance,cao2019learning,choudhuri2020partial,choudhuri2022coupling,choudhuri2023distribution}. A majority these methods are noise-sensitive, especially during the initial training stages, or are computationally expensive,  hindering the overall classification performance. Inspired by the works of Zhou and Dong et al. \cite{zhou2012ensemble,dong2020survey}, our approach employs ensemble learning to increase model robustness to noise by efficiently leveraging the diversity of data projections from multiple classifiers on target samples. Furthermore, we eliminate the need to access source data during the adaptation phase by inferring source prototypes before adaptation to the target domain, promoting source data privacy.

Model overfitting, particularly in noisy instances, is a prevalent challenge in deep neural networks, affecting task-specific performance. To combat this, various techniques are used to identify a subset of cleaner labels for training, like co-teaching frameworks \cite{han2018co}, meta-learning for weight estimation \cite{zhang2020distilling}, and Negative Learning \cite{kim2019nlnl}. However, these methods predominantly focus on random, uniformly distributed label noise, neglecting the distinct noise types that arise during domain shifts. This oversight makes them sensitive to thresholds, reducing their overall adaptability. Contrarily, our approach combines learning with complementary labels and an ensemble framework to generate confident target pseudo-labels, enabling better generalization across benchmarks.

Prior works on PDA \cite{cao2018partial,cao2018partial2,zhang2018importance,cao2019learning,choudhuri2020partial} have primarily concentrated on aligning domain distributions, overlooking the need for class-level distribution alignment. Models developed by Choudhuri et al. \cite{choudhuri2022coupling,choudhuri2023distribution} sought to align data distribution both within and across categories using the first-order moments of the distributions. However, they principally capture the distribution mean, overlooking the spread or variability of data points. Relying solely on first-order moments may be influenced by outliers, especially when the variability within categories is high between the source and target domains. In our study, we move beyond using first-order moments and add explicit objectives to ensure that data belonging to different categories fall under distinct class distributions and samples originating from the same class are aligned, regardless of their domains, resulting in more compact class distributions. To summarize, this work's primary contributions are as follows:

\begin{itemize}
\item Our approach integrates negative learning and ensemble frameworks in a PDA context, improving target predictions and preserving data privacy by eliminating the need for source data access during the adaptation phase.
\item We explore beyond first-order moments for distribution alignment in PDA and use explicit objectives to maximize inter-class separation and intra-class compactness.
\end{itemize}

\section{Related Work}

\noindent
Numerous studies have investigated modern domain adaptation techniques to minimize domain discrepancies and promote information transfer across domains, leveraging pre-existing labeled data \cite{hoffman2014lsda, yosinski2014transferable}. Many of these studies focus on either acquiring domain-invariant features or employing instance re-weighting schemes \cite{pan2010survey}. For instance, Ghifary et al. proposed a technique to minimize the differences in domain distributions while maintaining class distinguishability \cite{ghifary2016scatter}, and Pan et al. used the Maximum Mean Discrepancy metric in conjunction with a Transfer Component Analysis framework to decrease distribution disparities \cite{pan2010domain}. Similarly, Long et al.'s method \cite{long2013transfer} aims to align the marginal and conditional distributions of the domains. However, these approaches often fall short in complex adaptation tasks due to their dependence on shallow feature learning across domains.

Recent studies have addressed these limitations by employing deep learning frameworks to acquire complex, transferable features \cite{yosinski2014transferable,tzeng2014deep}. These studies generally aim to estimate and match distribution means in adaptation layers, with some leveraging the first-order MMD metric to create domain-invariant representations \cite{tzeng2014deep,zhang2018unsupervised}, while others merging adversarial loss with a domain classifier to provoke confusion and transform sample data in a domain-neutral way, as demonstrated in the works of Ganin et al. \cite{ganin2016domain} and Li et al.\cite{li2019joint}. However, the proposed networks are challenging to train and sensitive to hyper-parameters. They are typically confined to closed-set domain adaptation scenarios where the source and target label spaces are identical, limiting their scope in partial-domain adaptation contexts.

Relaxing the identical label set constraint of closed-set scenarios to encompass a larger source dataset often proves more pragmatic, enhancing the adaptation process by necessitating task-relevant information transfer from the source to the target. For example, the Selective Adversarial Network (SAN) lessens the weight of private source category samples using multiple adversarial networks to promote effective knowledge transfer in partial-domain adaptation scenarios \cite{cao2018partial,cao2018partial2}. Following this, newer works \cite{cao2018partial,cao2018partial2,zhang2018importance,cao2019learning} introduced frameworks for class-importance weight estimation and methods for measuring source domain samples' transferability, offering a softer metric to distinguish common categories from private source classes. However, early training stages in these models are highly sensitive to noisy feedback and can impede classification performance. Models addressing these issues \cite{choudhuri2022coupling,choudhuri2023distribution} align data distribution within and across categories using distribution means. However, these models overlook distribution variability and spread and introduce computational challenges due to the bottleneck of forward passing the entire source data in every epoch to estimate source prototypes. In this study, we aim to address these limitations.

\begin{figure*}[htbp!]
    \centering
    \includegraphics[width = 0.79\linewidth, height = 8.15 cm]{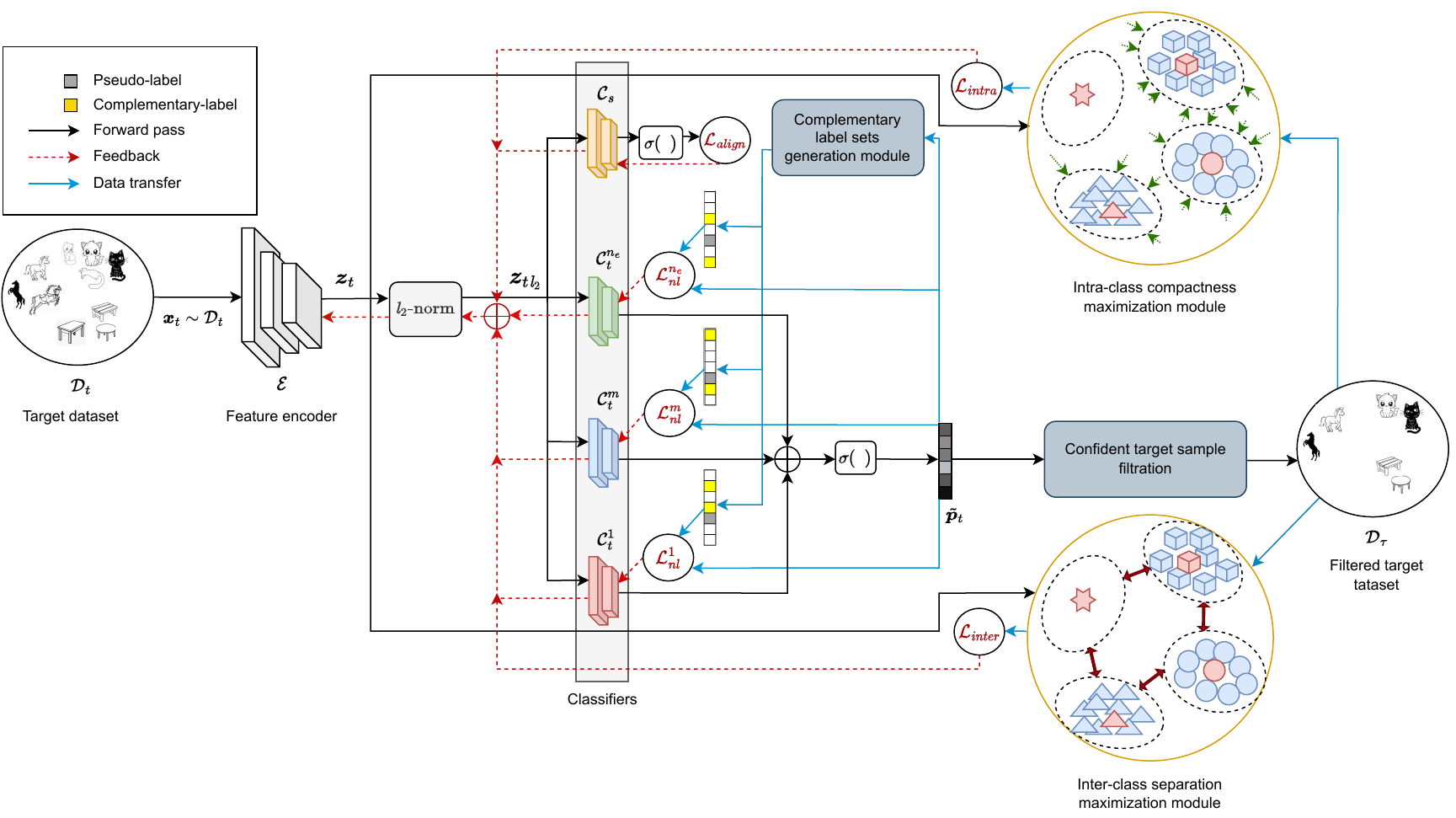}
    \caption{Architectural diagram of the proposed domain adaptation model (adaptation phase).}    
    \label{proposed_model}  
    \vspace{-5mm}
\end{figure*}

\section{Methodology}

\subsection{Problem Settings}
\label{prob_set}

\noindent
In this work, we explore a typical Partial Domain Adaptation (PDA) scenario, as put forth by Cao et al. \cite{cao2018partial}. The setup involves two distinct datasets from the source $s$ and the target $t$ domains. The source dataset $\mathcal{D}_s = \big\{(\boldsymbol{x}^i_s, y^i_s)\big\}_{i=1}^{n_s}$ encompasses $n_s$ labeled data points, $\boldsymbol{x}^i_s \in \mathbb{R}^{d_{\boldsymbol{x}}}$, sampled from a distribution $p_s$. A label $y^i_s$ belongs to a predefined label space $\mathcal{Y}_s$, containing $K_s$ distinct classes. In contrast, the target dataset $\mathcal{D}_t = \big\{\boldsymbol{x}^j_t\big\}_{j=1}^{n_t}$ includes $n_t$ unlabeled data points, $\boldsymbol{x}^j_t \in \mathbb{R}^{d_{\boldsymbol{x}}}$, sampled from distribution $p_t$. Furthermore, the label space of the target is considered to be contained within the source label space ($\mathcal{Y}_t \subseteq \mathcal{Y}_s$). It is worth noting that the target label space $\mathcal{Y}_t$ \textit{remains unknown during model training} and is only utilized for defining the PDA problem. Knowledge from the labeled data in $s$ is utilized to annotate a considerably smaller dataset $\mathcal{D}_t$. However, a domain shift is presumed to exist in a PDA setting such that $p_{s} \neq p_t$.  Similarly, a discrepancy exists between the distribution of the source samples with labels $y_s \in \mathcal{Y}_t$, denoted as $p_{s_{\mathcal{Y}_t}}$, and the target distribution ($p_{s_{\mathcal{Y}_t}} \neq p_t$). 

Given the task of classifying $\boldsymbol{x}_t \in \mathcal{D}_t$, our objective is to build a hypothesis classifier  $f \in \mathcal{H}$ ($\mathcal{H}$ - hypothesis space) that minimizes the \textit{target classification risk} $\epsilon_t(f)$. This is demonstrated in the following equation:

\vspace{-4mm}
\begin{equation}
    min \text{ } \epsilon_t(f), \text{ where } \epsilon_t(f) =  \mathbb{E}_{(\boldsymbol{x}_t,y_t) \sim {p_{t}}} \big[f(\boldsymbol{x}_t) \neq y_t\big]
\label{target_class_risk}
\end{equation}

\noindent
Leveraging the entire source domain data for estimating $f$ gives rise to the \textit{negative-transfer} problem; source samples with labels private to the source domain $\mathcal{Y}_s \setminus \mathcal{Y}_t$ ($\setminus$ denotes set-difference) contribute to an unwanted knowledge transfer, potentially increasing $\epsilon_t(f)$ in eq. \ref{target_class_risk}. Therefore, limiting the inclusion of these samples in the training phase is crucial to improving the classifier performance on $\mathcal{D}_t$.

\subsection{Proposed Approach}
\label{Proposed_Approach}

\noindent
We aim to approximate $f$ (see eq. \ref{target_class_risk}) by utilizing three families of networks: (a) a feature encoder $\mathcal{E}: \mathbb{R}^{d_{\boldsymbol{x}}} \rightarrow \mathbb{R}^{d_{\boldsymbol{z}}}$, parameterized by $\theta_{\mathcal{E}}$, that transforms a source/target input $\boldsymbol{x}$ into an encoded representation $\boldsymbol{z}$ \big($\boldsymbol{z} \in \mathbb{R}^{d_{\boldsymbol{z}}}$\big), (b) a source classifier $\mathcal{C}_s$, and (c) an ensemble of $n_e$ target classifiers $\{\mathcal{C}^m_t\}_{m=1}^{n_e}$, parameterized by $\theta_{\mathcal{C}_s}$ and $\{\theta_{\mathcal{C}^m_t}\}_{m=1}^{n_e}$, respectively (see fig. \ref{proposed_model}). The classifier networks transform the $\mathit{l}_2$-normalized encoded representation, $\boldsymbol{z}_{\mathit{l}_2}$ \big($\boldsymbol{z}_{\mathit{l}_2} := \frac{\boldsymbol{z}}{\|\boldsymbol{z}\|} \in [0,1]^{d_{\boldsymbol{z}}}$\big), into $K_s$ logits that are passed through a \textit{softmax} ($\sigma$) function to generate a $K_s$-dimensional probability vector, $\hat{\boldsymbol{p}} \in [0,1]^{K_s}$ \big($\mathcal{C}_s,  \{{\mathcal{C}^m_t}\}_{m=1}^{n_e}: [0,1]^{d_{\boldsymbol{z}}} \rightarrow [0,1]^{K_s}$\big). The negative transfer problem is alleviated by utilizing explicit objectives that encourage target sample alignment with source samples from shared categories. This is achieved by utilizing a subset of samples from $\mathcal{D}_t$, adaptively selected based on their prediction confidence and entropy. The method employs a robust pseudo-labeling method to improve target supervision, which fundamentally hinges on \textit{negative learning} \cite{kim2019nlnl}.

\subsubsection{Learning Source Category Prototypes}
\label{Learning_Source_Category_Prototypes}
We use class prototypes to align target features, offering computational efficiency over previous methods \cite{yue2021prototypical,liang2020we} that require computation of average latent features. This approach, inspired by Saito et al. \cite{saito2020universal}, robustly aligns features across two domains and ensures each class is represented during training updates. Furthermore, it eliminates the need for source data during the adaptation phase, which benefits data privacy. In our work, the neural network $\mathcal{C}_s$ consists of a linear layer with zero bias values. The weights in this layer, $\boldsymbol{\mu} = \big[\boldsymbol{\mu}_1, \boldsymbol{\mu}_2, \cdots, \boldsymbol{\mu}_{K_s}\big] \in \mathbb{R}^{d_{\boldsymbol{z}} \times K_s}$, can be interpreted as class prototypes. They are estimated by minimizing the categorical cross-entropy loss, as outlined below:

\vspace{-3mm}
\begin{equation}
    \mathcal{L}_{ce}\big(\theta_{\mathcal{C}_s},\theta_{\mathcal{E}}\big) =  
    - \frac{1}{n_s}  \sum\limits_{i=1}^{n_s} \sum\limits_{c=1}^{K_s} \boldsymbol{y}^i_{s,c} \text{ }log \text{ } {\hat{\boldsymbol{p}}^i_{s,c}}
\label{class}
\end{equation}

\begin{equation}
    {\hat{\boldsymbol{p}}^i_{s,c}} := \frac{exp\big(\boldsymbol{\mu}_{c}^T{\boldsymbol{z}^i_{s\hspace{0.25mm}{{\mathit{l}_2}}}}\big)}{\sum\limits_{c'=1}^{K_s}exp\big(\boldsymbol{\mu}_{c'}^T{\boldsymbol{z}^i_{s\hspace{0.25mm}{\mathit{l}_2}}}\big)}
\label{cross_entropy}
\end{equation}

\noindent
In eq. \ref{class} and \ref{cross_entropy}, $\boldsymbol{y}^i_{s}$ represents the one-hot encoded representation of label $y^i_s$. $\hat{\boldsymbol{p}}^i_{s}$ signifies the softmax output of $\mathcal{C}_s$. Subscript $c$ indexes the $c^{th}$ element of a vector.

The cross-entropy objective, commonly utilized for classification tasks \cite{zhang2018importance, ganin2016domain, cao2018partial,cao2018partial2}, focuses mainly on the ground-truth class, often neglecting crucial information from incorrect categories and not explicitly managing the inter-class separation. Inspired by Chen et al.'s research on complement objective training \cite{chen2019complement}, our method exploits this information from complement classes to reduce classifier uncertainty. We achieve this by averaging the sample-wise entropy over complement classes in a mini-batch and balancing their predicted probabilities through entropy maximization, thus reducing $\mathcal{L}_{comp}$ in eq. \ref{comp}. Our strategy emphasizes uncertain samples with higher confidence in reducing uncertainty (using $1-\hat{\boldsymbol{p}}_{g}$ in eq. \ref{comp2}, $g$ indexing the ground-truth entry). By normalizing the complement loss in eq. \ref{comp2} with the total number of complement categories, we ensure that the cross-entropy and complement objectives maintain the same scale.

\vspace{-4mm}
\begin{equation}
    \mathcal{L}_{comp}\big(\theta_{\mathcal{C}_s},\theta_{\mathcal{E}}\big) =  
    \frac{1}{n_s (K_s-1)} \sum\limits_{i=1}^{n_s} { l_{comp}}\Big(\hat{\boldsymbol{p}}^i_s, \boldsymbol{y}^i_s\Big)
\label{comp}
\end{equation}

\vspace{-3mm}
\begin{equation}
    l_{comp}(\hat{\boldsymbol{p}}^i_s,\boldsymbol{y}^i_s) := (1-\hat{\boldsymbol{p}}^i_{s,g}) \!\!\! \sum_{c = 1, c \neq g}^{K_s} \frac{\hat{\boldsymbol{p}}^i_{s,c}}{1-\hat{\boldsymbol{p}}^i_{s,g}} log \frac{\hat{\boldsymbol{p}}^i_{s,c}}{1-\hat{\boldsymbol{p}}^i_{s,g}}
    \label{comp2}
\end{equation}

Prior to the network's training using target samples, the class prototypes $[\boldsymbol{\mu}_c]_{c=1}^{K_s}$ (the weights of $\mathcal{C}_s$) are estimated by jointly training $\mathcal{E}$ and $\mathcal{C}$ on $\mathcal{D}_s$ while minimizing the two objectives, $\mathcal{L}_{ce}$ and $\mathcal{L}_{comp}$, as shown in eq. \ref{source_class_obj} ($\eta$ is a user-defined hyper-parameter and regulates the contribution of $\mathcal{L}_{comp}$ to the classification objective).

\vspace{-2mm}
\begin{equation}
    \min\limits_{\theta_{\mathcal{C}_s},\theta_{\mathcal{E}}} \Big\{\mathcal{L}_{ce}\big(\theta_{\mathcal{C}_s},\theta_{\mathcal{E}}\big) + \eta \hspace{0.5mm} \mathcal{L}_{comp}\big(\theta_{\mathcal{C}_s},\theta_{\mathcal{E}}\big)\Big\}
    \label{source_class_obj}
\end{equation}

\subsubsection{Aligning Target Samples with Source Prototypes}

Our objective is geared towards bringing the target samples nearer to their representative source prototypes $\boldsymbol{\mu} = [\boldsymbol{\mu}_c]_{c=1}^{K_s}$ (refer to sec. \ref{Learning_Source_Category_Prototypes}). However, prevalent domain shifts can cause the classifier to produce uniformly low probabilities across all categories, including the sample's true class, especially during the initial phases of adaptation. As a countermeasure, we train the joint networks $\mathcal{E}$ and $\mathcal{C}_s$ on $\mathcal{D}_t$ using the entropy minimization principle, as shown in eq. \ref{ent} (It is worth noting that the classifier weights (source prototypes) are not updated during this procedure). This encourages $\mathcal{E}$ to align the target features with the class prototypes.

\vspace{-1mm}
\begin{equation}
\label{ent}
    \mathcal{L}_{align}\big(\theta_{\mathcal{E}}\big) = 
    - \frac{1}{n_t}  \sum\limits_{j=1}^{n_t} \sum\limits_{c=1}^{K_s}
    {\hat{\boldsymbol{p}}^j_{t,c}} \text{ } log \text{ } \hat{\boldsymbol{p}}^j_{t,c}
\end{equation}
    
\begin{equation}
    \hat{\boldsymbol{p}}^j_{t,c} := \frac{exp\big(\boldsymbol{\mu}_{c}^T{\boldsymbol{z}^j_{t\hspace{0.25mm}{{\mathit{l}_2}}}}\big)}{\sum\limits_{c'=1}^{K_s}exp\big(\boldsymbol{\mu}_{c'}^T{\boldsymbol{z}^j_{t\hspace{0.25mm}{\mathit{l}_2}}}\big)}
\end{equation}

\subsubsection{Adaptive Target Supervision Using Pseudo-Labels}
\label{target_supervision}

Aligning target data with source prototypes via entropy minimization may cause mode-seeking behavior and overlook some class prototypes \cite{morerio2017minimal}. Furthermore, $\mathcal{E}$ is initially biased towards source data while generating class-discriminative features, and it retains information from private source categories $\mathcal{Y}_s \setminus \mathcal{Y}_t$. These factors can negatively impact the target classification performance. We employ a robust pseudo-labeling strategy to mitigate these challenges, leveraging negative learning via ensemble classifiers. The strategy refines target representations for better class-conditional distribution alignment. The pseudo-labeling framework comprises an ensemble network of $n_e$ target classifiers $\{\mathcal{C}^m_t\}_{m=1}^{n_e}$that are structurally identical to $\mathcal{C}_s$ (refer to sec. \ref{Proposed_Approach}). Each $\mathcal{C}^m_t$'s linear layer weights comprise learnable parameters $[\boldsymbol{w}^m_1, \boldsymbol{w}^m_2, \cdots, \boldsymbol{w}^m_{K_s} ] \in \mathbb{R}^{{d_z} \times K_s}$, initialized with the weights of $\mathcal{C}_s$. The prediction probability $\tilde{\boldsymbol{p}}^{j}_t(n)$ and pseudo-label $\tilde{y}^{j}_t(n)$ for the $n^{th}$ epoch are calculated using a moving average of $n_a$ previous ensemble output predictions, as shown below:

\vspace{-3mm}
\begin{equation}
\tilde{\boldsymbol{p}}^{j}_t(n) := {\sigma}\bigg(\frac{1}{n_a}\frac{1}{n_e}  \sum_{l=n-(n_a-1)}^{n} \sum_{m=1}^{n_e} {{\boldsymbol{w}}^{m}(l)}^T{\boldsymbol{z}^j_{t\hspace{0.25mm}{\mathit{l}_2}}}(l)\bigg)
\end{equation}

\begin{equation}
\label{hard_pseudo_labels}
    \tilde{y}^{j}_t(n) := argmax(\tilde{\boldsymbol{p}}^{j}_t(n))
\end{equation}

The target classifiers $\mathcal{C}^m_t$, initialized with source classifier weights, might underperform initially due to domain differences, leading to noisy target pseudo-labels. Conventional training methods that maximize the probability of a sample being categorized under its inferred pseudo-label can thus misguide the training process. To address this, we leverage negative learning \cite{kim2019nlnl}, which aims to lower the probability of incorrect label selection to $\frac{1}{K_s-1}$. The classifier is trained using a complementary label, assuming that the data sample does not belong to the complementary category. To enhance the robustness of pseudo-label refinement through the inclusion of diverse, complementary label feedback, we generate $n_e$ disjoint sets of complementary label indices (i.e., excluding the pseudo-label index) during each training epoch. "Each set, represented as $\{{cl}_m\}_{m=1}^{n_e}$ with $n_{cl}$ elements (shown in Algorithm \ref{algorithm_1}), is employed to train individual classifiers within the ensemble." 
For a certain complement category, the approach uses a weighting factor inversely proportional to its associated classification confidence ($1-\tilde{\boldsymbol{p}}^{j}_{t,c}$ in eq. \ref{negative_loss_ens}). The mechanism reduces the loss when the confidence is high, implying that the associated category is most likely the true label. Even if the correct label is mistakenly chosen as a complementary label, the effect of this incorrect feedback is alleviated by the presence of other complementary labels, thereby reducing pseudo-label generation noise.

\vspace{-3mm}
\begin{equation}
    \mathcal{L}_{nl}\big(\{\theta_{C^m_t}\}_{m=1}^{n_e},\theta_{\mathcal{E}}\big) = \frac{1}{n_e}\sum_{m=1}^{n_e}
    \mathcal{L}^m_{nl}(\theta_{\mathcal{C}^m_t},\theta_{\mathcal{E}},\mathcal{D}_t)
    \label{negative_loss_obj}
\end{equation}

\vspace{-6mm}
\begin{multline}
\mathcal{L}^m_{nl}\big(\theta_{\mathcal{C}^m_t},\theta_{\mathcal{E}}\big) =\\ - \frac{1}{n_t} \frac{1}{n_{cl}} \sum_{j=1}^{n_t}  \sum\limits_{c=1}^{K_s} (1-\tilde{\boldsymbol{p}}^{j}_{t,c})\hspace{0.5mm} \mathbbm{1}_{[c \in cl_m]} \hspace{0.5mm}log \hspace{0.5mm} {(1-\tilde{\boldsymbol{p}}^{j}_{t,c})}
\label{negative_loss_ens}
\end{multline}

Notably, during the initial few epochs, no pseudo-label refinement is implemented until the ensemble matures in its generalization capability. After multiple training epochs ($>$ 15), the increase in classification performance via the pseudo-label refinement process diminishes to a negligible level. Consequently, we employ standard supervised learning over \textit{highly-confident} samples using their pseudo-labels. This learning process involves a single model selected among $\mathcal{C}^m_t$ trained over the standard cross entropy loss to ensure a fair comparison with state-of-the-art approaches.

\begin{algorithm}
\caption{Complementary label sets generation}\label{alg:complementary}
\begin{algorithmic}[1]
\State \textbf{Input:} $\tilde{\boldsymbol{y}}^j_t, n_e, n_{cl}$ \Comment{$\tilde{\boldsymbol{y}}^j_t \rightarrow$ one-hot representation of $\tilde{{y}}^j_t$}
\State \textbf{Initialize:} $\boldsymbol{cl} = [\hspace{2mm}]$, $ind = \{\hspace{2mm}\}$, $cl_m = \{\hspace{2mm}\}$
\For{$1 \leq c \leq length(\tilde{\boldsymbol{y}}^j_t):$}
    \If{$\tilde{\boldsymbol{y}}^j_{t,c} = 0:$}
        \State $ind \gets ind \cup \{c\}$
    \EndIf
\EndFor
\For{$1 \leq m \leq n_e$:}
    \State $cl_m \gets random(ind,n_{cl})$ \Comment{$random(a,b)$ samples $b$ unique elements from $a$}
    \State $\boldsymbol{cl}[m] \gets cl_m$
    \State $ind \gets ind \setminus sub$  \Comment{$\setminus \rightarrow$ set-difference}
\EndFor
\State \textbf{Output:} $\boldsymbol{cl}$
\end{algorithmic}
\label{algorithm_1}
\end{algorithm}

\subsubsection{Filtering Confident Target Samples using CAC}

We aim to align the categorical distributions of samples by focusing on two primary goals: maximizing the distinction between different categories and enhancing cohesion within individual classes. This objective is realized by the use of pseudo-labels generated by ensemble classifiers. However, it's essential to note that not all pseudo-labels should be treated identically; Pseudo-labels associated with low confidence or high uncertainty can disrupt the classification process and divert it from its intended objective. To mitigate this issue, we strategically select a subset $\mathcal{D}_{\tau}$ of target samples from the dataset $\mathcal{D}_t$ that display above-average confidence. This selection process hinges on a dynamically estimated threshold parameter, $\tau$, determined by the predictive confidence and certainty as evaluated by the ensemble models. To provide further clarity, the high values of this metric are indicative of situations where the model exudes high confidence and/or certainty (manifested as low entropy). On the contrary, lower metric values suggest cases where the model lacks confidence or certainty (high entropy).

For each target sample $\boldsymbol{x}^j_t \in \mathcal{D}_t$, with an ensemble prediction vector of $\tilde{\boldsymbol{p}}^{j}_t$, we propose the ``Confidence-Adjusted Certainty" ($CAC$) metric as follows:

\begin{equation}
    CAC^j_t := 1 - \frac{H(\tilde{\boldsymbol{p}}^{j}_t)(1-max(\tilde{\boldsymbol{p}}^{j}_t))}{log_2 {K_s}}
    \label{cac_metric}
\end{equation}

\noindent
$H(\cdot)$ in eq. \ref{cac_metric} refers to Shannon's entropy. The metric amalgamates two crucial aspects of a classification model's performance: the inherent confidence in its predictions and the degree of uncertainty or variability associated with these predictions. Since the highest value for entropy is bounded by the $log$ of the number of categories, the metric is further normalized with $log_2 K_s$, with values spanning the interval [0,1], a higher value indicating a more confident target sample with low uncertainty in prediction.

Leveraging the pseudo-labels generated by the ensemble model, we compute the \textit{average} $CAC$ over samples in $\mathcal{D}_t$ and assign this as the threshold $\tau$. For all the target samples whose $CAC$ value exceeds $\tau$, are included in the refined dataset, $\mathcal{D}_{\tau}$.

\vspace{2mm}
\subsubsection{Maximizing Inter-Class Separation}
\label{sec_inter}

We employ $\mathcal{D}_{\tau}$ to steer the learning process towards aligning class-conditional distributions. The key idea is to ensure that data belonging to different categories fall under distinct class distributions, regardless of their domains. To achieve this, our strategy minimizes the inter-class objective $\mathcal{L}_{inter}$ (eq. \ref{inter}) to widen the gap between (a) samples within the target domain and (b) between target samples and source prototypes, \textit{under the condition that they are from different categories} ($\tilde{y}_t$, in eq. \ref{inter}, signifies the associated pseudo-label category index of a sample $\boldsymbol{x}_t$ (see eq. \ref{hard_pseudo_labels}). $c$ represents the category index of source prototype $\boldsymbol{\mu}_c$ ($1 \leq c \leq K_s)$.

\vspace{-5mm}
\begin{multline}
\label{inter}
    \mathcal{L}_{inter}(\theta_{\mathcal{E}}) = - \Vast(
    \cfrac{\sum\limits_{\substack{(\boldsymbol{x}^i_t,\tilde{y}^i_t), 
    (\boldsymbol{x}^j_t,\tilde{y}^j_t) \in \mathcal{D}_{\tau}\\i \neq j}} \!\!\!\!\!\!\! \mathbbm{1}_{[\tilde{y}^i_t \neq \tilde{y}^j_t]} \text{ } \delta(\boldsymbol{z}^i_{t},\boldsymbol{z}^j_{t})}{\sum\limits_{\substack{(\boldsymbol{x}^i_t,\tilde{y}^i_t), (\boldsymbol{x}^j_t,\tilde{y}^j_t) \in \mathcal{D}_{\tau}\\i \neq j}} \!\!\!\!\!\!\! \mathbbm{1}_{[\tilde{y}^i_t \neq \tilde{y}^j_t]}} \text{ }+\\ 
    \cfrac{\sum\limits_{(\boldsymbol{x}_t,\tilde{y}_t) \in \mathcal{D}_{\tau}} \sum\limits_{\boldsymbol{\mu}_c \in \boldsymbol{\mu}} \mathbbm{1}_{[\tilde{y}_t \neq c]} \text{ } \delta(\boldsymbol{z}_{t},\boldsymbol{\mu}_c)}{\sum\limits_{(\boldsymbol{x}_t,\tilde{y}_t) \in \mathcal{D}_{\tau}} \sum\limits_{\boldsymbol{\mu}_c \in \boldsymbol{\mu}} \mathbbm{1}_{[\tilde{y}_t \neq c]}} \Vast)
\end{multline}

\vspace{-5mm}
\begin{equation}
\label{dist_func}
    \delta(\boldsymbol{z},\boldsymbol{z}') := 1 - \frac{\boldsymbol{z} \cdot \boldsymbol{z}'}{\| \boldsymbol{z} \| \cdot \| \boldsymbol{z}'\| }
\end{equation}

\noindent

\subsubsection{Maximizing Within-Class Compactness}
\label{sec_intra}

In this section, we detail an objective designed to align samples originating from the same class, which results in more compact class distributions. This is realized by minimizing the distance between the latent representations of any pair of samples falling under the same category, regardless of the domains they originate from. The subsequent intra-class objective $\mathcal{L}_{intra}$ is represented by the following equation:

\vspace{-5mm}
\begin{multline}
    \mathcal{L}_{intra}(\theta_{\mathcal{E}}) =
     \cfrac{\sum\limits_{\substack{(\boldsymbol{x}^i_t,\tilde{y}^i_t), 
    (\boldsymbol{x}^j_t,\tilde{y}^j_t) \in \mathcal{D}_{\tau}\\i \neq j}} \!\!\!\!\!\!\! \mathbbm{1}_{[\tilde{y}^i_t = \tilde{y}^j_t]} \text{ } \delta(\boldsymbol{z}^i_{t},\boldsymbol{z}^j_{t})}{\sum\limits_{\substack{(\boldsymbol{x}^i_t,\tilde{y}^i_t), (\boldsymbol{x}^j_t,\tilde{y}^j_t) \in \mathcal{D}_{\tau}\\i \neq j}} \!\!\!\!\!\!\! \mathbbm{1}_{[\tilde{y}^i_t = \tilde{y}^j_t]}} \text{ }+\\ 
    \cfrac{\sum\limits_{(\boldsymbol{x}_t,\tilde{y}_t) \in \mathcal{D}_{\tau}} \sum\limits_{\boldsymbol{\mu}_c \in \boldsymbol{\mu}} \mathbbm{1}_{[\tilde{y}_t = c]} \text{ } \delta(\boldsymbol{z}_{t},\boldsymbol{\mu}_c)}{\sum\limits_{(\boldsymbol{x}_t,\tilde{y}_t) \in \mathcal{D}_{\tau}} \sum\limits_{\boldsymbol{\mu}_c \in \boldsymbol{\mu}} \mathbbm{1}_{[\tilde{y}_t = c]}}
\label{intra}
 \end{multline}  
     
\subsubsection{Overall Objective}

The comprehensive objective  for extracting target labels can be summarized as follows (with $\alpha$ and $\beta$ being user-defined hyper-parameters that determine the contribution of each objective in the learning process):

\vspace{-3mm}
\begin{multline}
\label{overall_obj}
\min\limits_{(\theta_{\mathcal{C}},\theta_{\mathcal{E}})} \Big\{
\mathcal{L}_{nl}\big(\{\theta_{C^m_t}\}_{m=1}^{n_e},\theta_{\mathcal{E}}\big) 
+  \alpha\mathcal{L}_{inter}({\theta_{\mathcal{E}}}) + \\
\beta \mathcal{L}_{intra}({\theta_{\mathcal{E}}}) + \mathcal{L}_{align}(\theta_{\mathcal{E}}) \Big\}
\end{multline}

\section{Experiments}

\noindent
In this section, we present our comprehensive evaluation of the proposed model against the current state-of-the-art techniques using three benchmark datasets for domain adaptation. Our evaluation covers a wide range of PDA settings, with multiple adaptation tasks to ensure a thorough assessment. In line with the standard evaluation criteria \cite{cao2018partial,cao2018partial2,tzeng2017adversarial}, we use classification accuracy as the comparison metric and include all labeled source data and unlabeled target data for Partial Domain Adaptation. Furthermore, we present a comprehensive analysis of the model performance, including the effectiveness of \textit{ensemble learning}, \textit{target supervision using confident samples}, \textit{intra/inter-class distribution optimization}, and \textit{target supervision using complementary label sets}. In the following sections, we present the results of our experiments and an ablation analysis of the mentioned modules.

\subsection{Datasets} 
\label{datasets}

\noindent
To evaluate the transferability of domain information and the accuracy of target classification, we employ three commonly used image datasets for domain adaptation: \textit{Office-31} \cite{saenko2010adapting}, \textit{Office-Home} \cite{venkateswara2017deep}, and \textit{VisDA 2017} \cite{peng2017visda}.\\

\textbf{Office-31:} The Office-31 dataset \cite{saenko2010adapting} is composed of 4652 RGB images from three distinct domains: Amazon (A), DSLR (D), and Webcam (W). These images are classified into 31 categories. To establish a Partial Domain Adaptation setup, we adopt the standard protocol proposed by Cao et al. \cite{cao2018partial}, where the target dataset includes samples from 10 categories. To conduct a thorough evaluation, we assess the proposed model for multiple adaptation tasks on the following source-target domain pairs: A$\rightarrow$D, A$\rightarrow$W, D$\rightarrow$A, D$\rightarrow$W, W$\rightarrow$A, and W$\rightarrow$D.\\

\begin{table*}[htbp!]
\centering
\begin{adjustbox}{width=0.70\textwidth}
\begin{tabular}{>{\bfseries}c || c c c c c c | c || c c | c}
\hline
\multirow{2}{*}{Method} & \multicolumn{7}{c||}{\textbf{Office-31}} & \multicolumn{3}{c}{\textbf{VisDA 2017}} \\
\cline{2-11}
& \textbf{A $\rightarrow$ D} & \textbf{A $\rightarrow$ W} & \textbf{D $\rightarrow$ A} & \textbf{D $\rightarrow$ W} & \textbf{W $\rightarrow$ A} & \textbf{W $\rightarrow$ D} & \textbf{Avg.} & \textbf{R $\rightarrow$ S} & \textbf{S $\rightarrow$ R} & \textbf{Avg.} \\
\hline
Resnet-50\cite{he2016deep} & 83.44 & 75.59 & 83.92 & 96.27 & 84.97 & 98.09 & 87.05 & 64.30 & 45.30 & 54.80 \\
\hline
DANN\cite{ganin2016domain} & 81.53 & 73.56 & 82.78 & 96.27 & 86.12 & 98.73 & 86.50 & 73.84 & 51.01 & 62.43 \\
ADDA\cite{tzeng2017adversarial} & 83.41 & 75.67 & 83.62 & 95.38 & 84.25 & 99.85 & 87.03 & - & - & -\\
\hline
PADA\cite{cao2018partial2} & 82.17 & 86.54 & 92.69 & 99.32 & 95.41 & {\color{red}\textbf{100.00}} & 92.69 & 76.50 & 53.50 & 65.00 \\
DRCN\cite{li2020deep} & 88.50 & {\color{red}\textbf{100.00}} & {\color{red}\textbf{100.00}} & 86.00 & 95.60 & 95.80 & 94.30 & 74.20 & 57.20 & 65.70 \\
IWAN\cite{zhang2018importance} & 90.45 & 89.15 & 95.62 & 99.32 & 94.26 & 99.36 & 94.69 & 71.30 & 48.60 & 59.95 \\
SAN\cite{cao2018partial} & 94.27 & 93.90 & 94.15 & 99.32 & 88.73 & 99.36 & 94.96 & 69.70 & 49.90 & 59.80 \\
ETN\cite{cao2019learning} & 95.03 & 94.52 & 96.21 & {\color{red}\textbf{100.00}} & 94.64 & {\color{red}\textbf{100.00}} & 96.73 & 78.24 & 68.53 & 73.39 \\
SRL\cite{choudhuri2020partial} & 94.46 & 92.07 & 93.68 & 95.84 & 93.72 & 99.24 & 94.84 & 73.96 & 54.12 & 64.04 \\
\hline
Proposed Model & {\color{red}\textbf{98.20}} & 98.46 & 96.12 & {\color{red}\textbf{100.00}} & {\color{red}\textbf{95.68}} & {\color{red}\textbf{100.00}} & {\color{red}\textbf{98.08}} & {\color{red}\textbf{78.41}} & {\color{red}\textbf{74.27}} & {\color{red}\textbf{76.34}} \\
\hline
\end{tabular}
\end{adjustbox}
\vspace{1mm}
\caption{Accuracy of classification (\%) for PDA tasks on the Office-31 and Visda 2017 datasets (Resnet-50 backbone)}
\vspace{-3mm}
\label{office-31}
\end{table*}

\begin{table*}[htbp!]
\centering
\begin{adjustbox}{width=1\textwidth}
\begin{tabular}{>{\bfseries} c || c c c c c c c c c c c c | c}
 \hline
Method & \textbf{Ar $\rightarrow$ Cl} & \textbf{Ar $\rightarrow$ Pr} & \textbf{Ar $\rightarrow$ Rw} & \textbf{Cl $\rightarrow$ Ar} & \textbf{Cl $\rightarrow$ Pr} & \textbf{Cl $\rightarrow$ Rw} & \textbf{Pr $\rightarrow$ Ar} & \textbf{Pr $\rightarrow$ Cl} & \textbf{Pr $\rightarrow$ Rw} & \textbf{Rw $\rightarrow$ Ar} & \textbf{Rw $\rightarrow$ Cl} & \textbf{Rw $\rightarrow$ Pr} & \textbf{Avg.}\\ 
 \hline
Resnet-50\cite{he2016deep} & 46.33 & 67.51 & 75.87 & 59.14 & 59.94 & 62.73 & 58.22 & 41.79 & 74.88 & 67.40 & 48.18 & 74.17 & 61.35\\
\hline
DANN\cite{ganin2016domain} & 43.76 & 67.90 & 77.47 & 63.73 & 58.99 & 67.59 & 56.84 & 37.07 & 76.37 & 69.15 & 44.30 & 77.48 & 61.72\\
ADDA\cite{tzeng2017adversarial} & 45.23 & 68.79 & 79.21 & 64.56 & 60.01 & 68.29 & 57.56 & 38.89 & 77.45 & 70.28 & 45.23 & 78.32 & 62.82\\
\hline
PADA\cite{cao2018partial2} & 51.95 & 67.00 & 78.74 & 52.16 & 53.78 & 59.03 & 52.61 & 43.22 & 78.79 & 73.73 & 56.60 & 77.09 & 62.06\\
DRCN\cite{li2020deep} & 54.00 & 76.40 & 83.00 & 62.10 & 64.50 & 71.00 & 70.80 & 49.80 & 80.50 & 77.50 & 59.10 & 79.90 & 69.00\\
IWAN\cite{zhang2018importance} & 53.94 & 54.45 & 78.12 & 61.31 & 47.95 & 63.32 & 54.17 & 52.02 & 81.28 & 76.46 & 56.75 & 82.90 & 63.56\\
SAN\cite{cao2018partial} & 44.42 & 68.68 & 74.60 & 67.49 & 64.99 & {\color{red}\textbf{77.80}} & 59.78 & 44.72 & 80.07 & 72.18 & 50.21 & 78.66 & 65.30\\
ETN\cite{cao2019learning} & 59.24 & 77.03 & 79.54 & 62.92 & 65.73 & 75.01 & 68.29 & {\color{red}\textbf{55.37}} & 84.37 & 75.72 & 57.66 & {\color{red}\textbf{84.54}} & 70.45\\
SRL\cite{choudhuri2020partial} & 56.21 & 73.34 & 80.63 & 64.08 & 61.72 & 66.41 & 70.83 & 53.13 & 83.57 & 77.01 & {\color{red}\textbf{58.31}} & 81.24 & 68.87\\
 \hline
Proposed Model & {\color{red}\textbf{61.03}} & {\color{red}\textbf{84.03}} & {\color{red}\textbf{90.10}} & {\color{red}\textbf{70.21}} & {\color{red}\textbf{74.60}} & 77.46 & {\color{red}\textbf{71.05}} & 55.21 & {\color{red}\textbf{86.36}} & {\color{red}\textbf{79.16}} & 58.15 & 84.13 & {\color{red}\textbf{74.29}}\\
\hline
\end{tabular}
\end{adjustbox}
\vspace{0.5mm}
\caption{Accuracy of classification (\%) for PDA tasks on the Office-Home dataset (Resnet-50 backbone)}
\vspace{-7mm}
\label{office-home}
\end{table*}

\textbf{Office-Home:} Office-Home \cite{venkateswara2017deep} is a larger dataset that comprises 15,500 RGB images from four domains, namely Artistic (Ar), Clip Art (Cl), Product (Pr), and Real-world (Rw). In line with the evaluation setup presented for Office-31, we follow the same protocol and create the source and target datasets with 65 and 25 categories, respectively. To conduct a thorough evaluation, we consider 12 different adaptation tasks, namely Ar$\rightarrow$Cl, Ar$\rightarrow$Pr, Ar$\rightarrow$Rw, Cl$\rightarrow$Ar, Cl$\rightarrow$Pr, Cl$\rightarrow$Rw, Pr$\rightarrow$Ar, Pr$\rightarrow$Cl, Pr$\rightarrow$Rw, Rw$\rightarrow$Ar, Rw$\rightarrow$Cl, and Rw$\rightarrow$Pr.
\vspace{2mm}

\textbf{VisDA 2017}: VisDA 2017 \cite{peng2017visda} is a robust dataset designed to evaluate domain adaptation models. With a total of 207,785 images across 12 distinct categories, it is divided into two primary domains: Synthetic images (S), made up of 2D renderings generated from 3D models from varied perspectives, and Real images (R) that feature photo-realistic pictures. Given the provided domains, two cross-domain learning tasks have been constructed: S$\rightarrow$R and R$\rightarrow$S.

\begin{table}[htbp!]
\centering
\begin{tabular}{>{\bfseries} c || c c c c c c}
  \hline
 Datasets & $ \hspace{2mm} \boldsymbol{n_a} \hspace{2mm}$ & $\hspace{2mm} \boldsymbol{n_e} \hspace{2mm}$ & $\hspace{2mm} \boldsymbol{n_{cl}} \hspace{2mm}$ & $\hspace{2mm} \boldsymbol{\alpha} \hspace{2mm}$ & $\hspace{2mm} \boldsymbol{\beta} \hspace{2mm}$ & $\hspace{2mm} \boldsymbol{\eta}$\\
  \hline
 Office-31 &10 & 3 & 3 & 0.5 & 1.5 & 1.5\\
 Office-home & 10 & 3 & 3 & 0.7 & 1.9 & 1.1\\
 VisDA 2017 & 10 & 3 & 3 & 0.7 & 1.9 & 1.1\\
  \hline
\end{tabular}
\vspace{1.5mm}
\caption{Parameter settings for model evaluation.}
\label{parameter_settings}
\end{table}

\vspace{-7mm}
\subsection{Implementation}

\noindent
Our experiments used models implemented in PyTorch on an Nvidia 3090-Ti GPU with 24 GB memory. For encoding the source/target samples, the backbone of our structure incorporated the Resnet-50 model, pre-trained on the ImageNet dataset. We built the feature encoder atop this backbone network, denoted as $\mathcal{E}$. This was achieved by omitting the last dense layer. Additionally, we modified the network by eliminating its last linear layer and incorporating a randomly initialized weight matrix $\boldsymbol{\mu}$ as the source classifier, $\mathcal{C}_s$. The process of fine-tuning the model was performed on source samples. The learning rate for the linear layers was set at 0.001, a value ten times greater than the learning rate for the feature encoder. Regarding the learning rate schedule, we followed the formula $lr(n) = lr(0)(1 + \gamma_{lr}\cdot n)^{- \alpha_{lr}}$, where $lr(0)$ represents the initial learning rate. Here, we established $\eta_0$ as 0.01, $\gamma_{lr}$ as 0.0002, and $\alpha_{lr}$ as 0.75. The training was conducted using mini-batch Stochastic Gradient Descent (SGD) with a momentum parameter set to 0.9. For a total of 250 epochs, $\mathcal{E}$ and $\mathcal{C}_s$ were trained on the source samples. Target classifiers mirror $\mathcal{C}_s$'s structure and initialize using its learned weights. The batch size for source and target data was set to 32 during adaptation. The ensemble model was trained for 2500 epochs. The parameter settings for our experiments are displayed in Table \ref{parameter_settings}. 
During model evaluation, one classifier from $\mathcal{C}^m_t$ is chosen for target prediction to ensure a fair comparison with leading methods.

\begin{figure*}[htbp!]
    \centering
    \begin{minipage}{0.42\textwidth}
        \centering
        \includegraphics[width=1.0\linewidth]{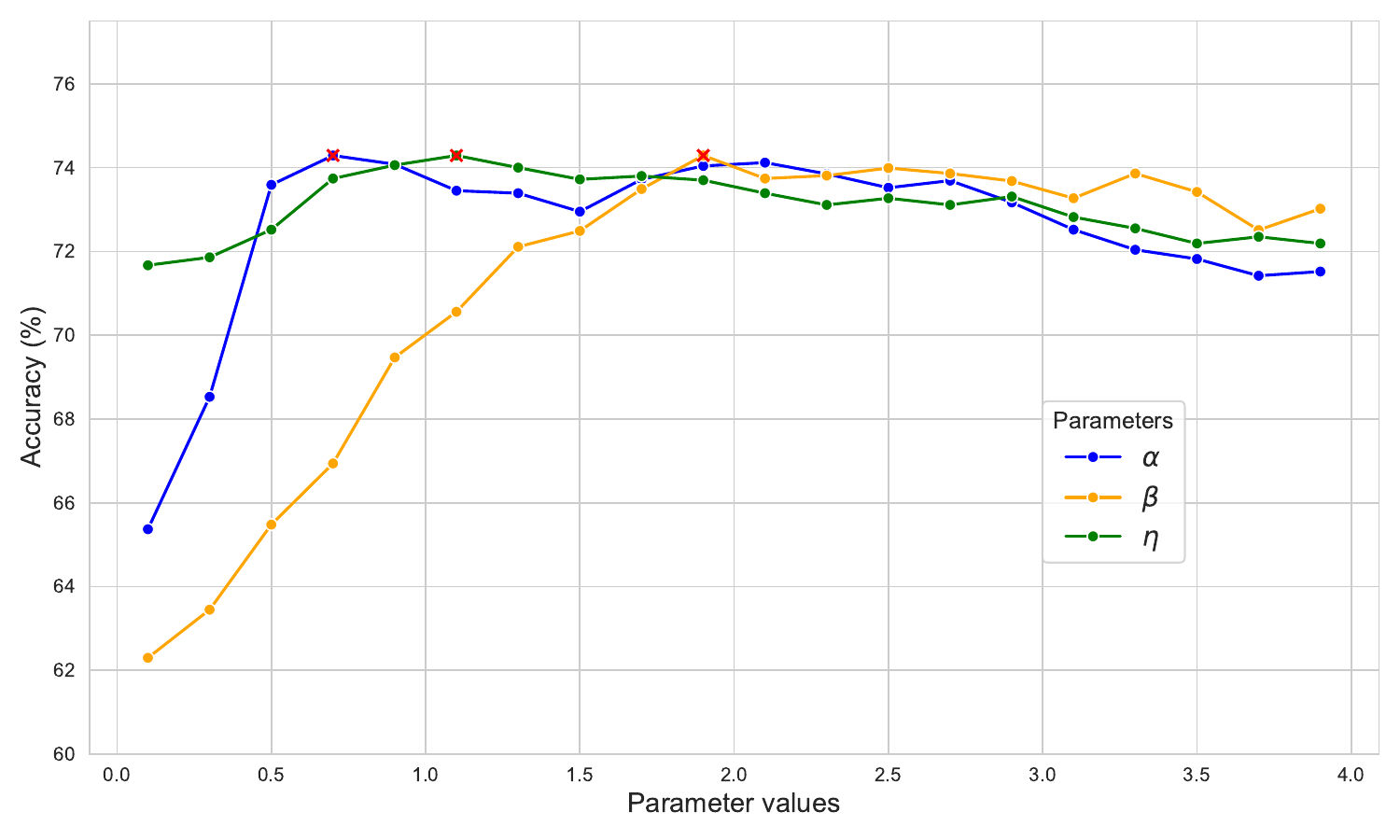} 
    \end{minipage}\hfill 
    \begin{minipage}{0.58\textwidth}
        \centering
        \includegraphics[width=1.0\linewidth]{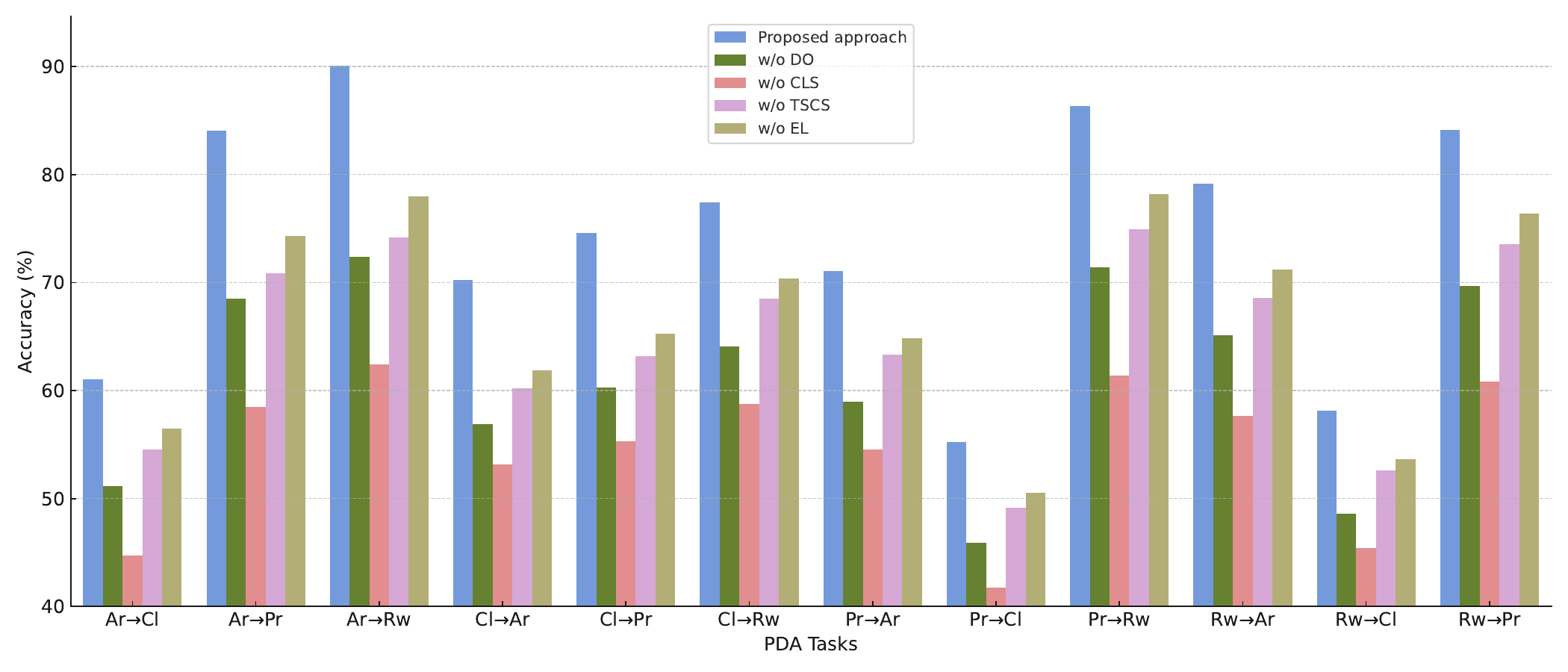} 
    \end{minipage}
    \vspace{-4mm}
    \caption{(a) Sensitivity analysis of $\alpha$, $\beta$, and $\eta$, and (b) Accuracy results of ablation analysis on the Office-Home dataset.}
    \label{plot_ablation1} 
    \vspace{-5mm}
     
\end{figure*}

\vspace{-0.1cm}
\subsection{Comparison Models}

\noindent
We use the target classification accuracy metric to assess our method against leading models for partial-domain adaptation. The models considered for comparison include a variety of state-of-the-art networks. These encompass the Domain Adversarial Neural Network (DANN) \cite{ganin2016domain}, Partial Adversarial Domain Adaptation (PADA) \cite{cao2018partial}, Adversarial Discriminative Domain Adaptation (ADDA) network \cite{tzeng2017adversarial}, Importance Weighted Adversarial Nets (IWAN) \cite{zhang2018importance}, Example Transfer Network (ETN) \cite{cao2019learning}, Selective Adversarial Network (SAN) \cite{cao2018partial}, Deep Residual Correction Network (DRCN) \cite{li2020deep}, and Selective Representation Learning For Class-Weight Computation (SRL) \cite{choudhuri2020partial}. To highlight the problem of negative transfer present in the DANN and ADDA models (which are designed to address closed-set adaptation tasks), we include the classification accuracy of Resnet-50 \cite{he2016deep}. This is done by training the model solely on the target data in a supervised manner to provide a meaningful benchmark for comparison.

\vspace{-0.1cm}
\subsection{Classification Results}

\noindent
The target classification accuracies on Office-31, VisDA 2017, and Office-Home benchmark datasets are presented in tables \ref{office-31} and \ref{office-home}, respectively. It is noteworthy that the accuracy values for Resnet-50 \cite{he2016deep} and DANN \cite{ganin2016domain} in tasks A $\rightarrow$ W, A $\rightarrow$ D, D $\rightarrow$ A (Table \ref{office-31}) and Ar $\rightarrow$ Cl, Cl $\rightarrow$ Pr, Pr $\rightarrow$ Ar, Pr $\rightarrow$ Cl, and Rw $\rightarrow$ Cl (Table \ref{office-home}) indicate the existence of the negative transfer problem; the DANN model, designed for closed-set domain adaptation, fails to filter out the impact of samples from the private source domain classes ($\mathcal{Y}_s - \mathcal{Y}_t$), thereby impacting its accuracy. 

Empirical evidence, as displayed in Tables \ref{office-31} and \ref{office-home}, testifies to the superior performance of our proposed model. Notably, it records the highest classification accuracies in four out of six tasks and eight out of twelve tasks on the respective datasets. Furthermore, it delivers the \textit{highest average accuracies} across both datasets, further underlining its efficacy.

\subsection{Parameter Sensitivity}
\label{param_sens}

\noindent
The trade-off parameters $\alpha$, $\beta$, and $\eta$ play pivotal roles in the training process. The first two regulate the influence of the inter-class and inter-category distribution alignment objectives, while $\eta$ controls the complementary entropy training on the source samples. Figure \ref{plot_ablation1} (a) demonstrates how these parameters influence the target classification accuracy on the Office-Home dataset. It's observed that when there's an increase in a particular hyperparameter's value, the accuracy fluctuates within a tight margin of 3.5\% from the peak value. This stability suggests that our proposed methodology exhibits resilience against variations in these parameters.

\subsection{Ablation Analysis}

\noindent
In this study, we postulate the importance of the following elements within our suggested network: (a) Ensemble Learning (EL), (b) Target Supervision Using Confident Samples (TSCS), (c) Ensemble Learning from Complementary Label Sets (CLS), and (d) Intra/Inter-Category Distribution Optimization (DO). To evaluate the impact of these components, an ablation analysis is carried out on twelve PDA tasks using the Office-Home dataset, wherein each component is individually deactivated to assess its effect on performance and the necessity for target accuracy enhancement.
\vspace{2mm}

\begin{itemize}
    \item \textbf{w/o EL}: To understand the role of ensemble learning, we adjust the $n_e$ value to 1, which minimizes the number of target ensemble classifiers. As displayed in Figure \ref{plot_ablation1} (b), there is a consistent reduction in classification accuracy compared to our proposed model.
    \vspace{2mm}
    
    \item \textbf{w/o TSCS}: Our hypothesis posits that not all ensemble-generated pseudo-labels are of equal value. Low-confidence labels can misguide the classification process. 
    To gauge the impact, we eliminate the creation of  $\mathcal{D}_{\tau}$ and instead use all the target samples for class-level distribution alignment. A steep decline in the network performance, as shown in Figure \ref{plot_ablation1} (b), underscores the significance of the TSCS module.
    
    \vspace{2mm}
    \item \textbf{w/o CLS}: The use of complementary label sets is intended to improve the robustness of the pseudo-label refinement process by incorporating diverse, complementary label feedback. To evaluate this, we restrict the $n_{cl}$ value to 1, consequently limiting the diversity of ensemble training through different complementary labels by generating a single set of complementary label index set $cl$ that is shared among ensemble models. As shown in figure \ref{plot_ablation1} (b), this leads to a significant performance decline (most pronounced performance decrease across tasks), thus substantiating the necessity for diversification. 
    \vspace{0mm}

    \item \textbf{w/o DO}: The target classifiers, initialized with source classifier weights, may not initially provide optimal target classification performance due to existing domain discrepancy. To overcome this, we introduce the $\mathcal{L}_{inter}$ and $\mathcal{L}_{intra}$ objectives to maximize inter-category distance and to improve class compactness in the latent space using information beyond the first-order moments of the distributions, in a domain-invariant fashion. To evaluate their influence, we set $\alpha$ and $\beta$ to 0. As demonstrated in figure \ref{plot_ablation1} (b), there is a noticeable decrease in average classification accuracy across all tasks.
\end{itemize}

\vspace{1mm}
\section{Conclusion}

\noindent
In summary, our research introduces a robust Partial Domain Adaptation (PDA) framework designed to counter the issue of negative transfer through a robust target-supervision strategy. Our approach uniquely incorporates ensemble learning on negative classes to enhance pseudo-label refinement. 
We look beyond traditional PDA techniques that rely on first-order moments to optimize intra-class compactness and inter-class separation using source prototypes and confident target samples. Additionally, our framework ensures data privacy, eliminating the requirement for source data during adaptation. Extensive tests across various adaptation tasks underscore our framework's robustness and superior performance over existing PDA approaches.

\printbibliography

@inproceedings{liang2020we,
  title={Do we really need to access the source data? source hypothesis transfer for unsupervised domain adaptation},
  author={Liang, Jian and Hu, Dapeng and Feng, Jiashi},
  booktitle={International Conference on Machine Learning},
  pages={6028--6039},
  year={2020},
  organization={PMLR}
}

@article{pan2010domain,
  title={Domain adaptation via transfer component analysis},
  author={Pan, Sinno Jialin and Tsang, Ivor W and Kwok, James T and Yang, Qiang},
  journal={IEEE transactions on neural networks},
  volume={22},
  number={2},
  pages={199--210},
  year={2010},
  publisher={IEEE}
}

@inproceedings{saenko2010adapting,
  title={Adapting visual category models to new domains},
  author={Saenko, Kate and Kulis, Brian and Fritz, Mario and Darrell, Trevor},
  booktitle={European conference on computer vision},
  pages={213--226},
  year={2010},
  organization={Springer}
}

@inproceedings{torralba2011unbiased,
  title={Unbiased look at dataset bias},
  author={Torralba, Antonio and Efros, Alexei A},
  booktitle={CVPR 2011},
  pages={1521--1528},
  year={2011},
  organization={IEEE}
}

@inproceedings{cao2018partial,
  title={Partial transfer learning with selective adversarial networks},
  author={Cao, Zhangjie and Long, Mingsheng and Wang, Jianmin and Jordan, Michael I},
  booktitle={Proceedings of the IEEE conference on computer vision and pattern recognition},
  pages={2724--2732},
  year={2018}
}

@inproceedings{cao2018partial2,
  title={Partial adversarial domain adaptation},
  author={Cao, Zhangjie and Ma, Lijia and Long, Mingsheng and Wang, Jianmin},
  booktitle={Proceedings of the European Conference on Computer Vision (ECCV)},
  pages={135--150},
  year={2018}
}

@inproceedings{zhang2018importance,
  title={Importance weighted adversarial nets for partial domain adaptation},
  author={Zhang, Jing and Ding, Zewei and Li, Wanqing and Ogunbona, Philip},
  booktitle={Proceedings of the IEEE conference on computer vision and pattern recognition},
  pages={8156--8164},
  year={2018}
}

@article{ganin2016domain,
  title={Domain-adversarial training of neural networks},
  author={Ganin, Yaroslav and Ustinova, Evgeniya and Ajakan, Hana and Germain, Pascal and Larochelle, Hugo and Laviolette, Fran{\c{c}}ois and Marchand, Mario and Lempitsky, Victor},
  journal={The journal of machine learning research},
  volume={17},
  number={1},
  pages={2096--2030},
  year={2016},
  publisher={JMLR. org}
}

@INPROCEEDINGS{1,
title={Partially adversarial learning and adaptation},
author={Chien, Jen Tzung and Lyu, Yu Ying},
booktitle={2019 27th European Signal Processing Conference (EUSIPCO)},
pages={1--5},
year={2019},
organization={IEEE}
}

@ARTICLE{2,
  author={P. {Li} and D. {Zhang} and P. {Chen} and X. {Liu} and A. {Wulamu}},
  journal={IEEE Access}, 
  title={Multi-Adversarial Partial Transfer Learning With Object-Level Attention Mechanism for Unsupervised Remote Sensing Scene Classification}, 
  year={2020},
  volume={8},
  number={},
  pages={56650-56665},
  doi={10.1109/ACCESS.2020.2982034}}

@inproceedings{3,
  title={Learning to transfer examples for partial domain adaptation},
  author={Cao, Zhangjie and You, Kaichao and Long, Mingsheng and Wang, Jianmin and Yang, Qiang},
  booktitle={Proceedings of the IEEE Conference on Computer Vision and Pattern Recognition},
  pages={2985--2994},
  year={2019}
}

@ARTICLE{4,
  author={C. -X. {Ren} and P. {Ge} and P. {Yang} and S. {Yan}},
  journal={IEEE Transactions on Neural Networks and Learning Systems},
  title={Learning Target-Domain-Specific Classifier for Partial Domain Adaptation}, 
  year={2020},
  volume={},
  number={},
  pages={1-13},
  doi={10.1109/TNNLS.2020.2995648}}

@inproceedings{5,
  title={Multi-Weight Partial Domain Adaptation.},
  author={Hu, Jian and Tuo, Hongya and Wang, Chao and Qiao, Lingfeng and Zhong, Haowen and Jing, Zhongliang},
  booktitle={BMVC},
  pages={5},
  year={2019}
}

@ARTICLE{7,
  author={J. {Chen} and X. {Wu} and L. {Duan} and S. {Gao}},
  journal={IEEE Transactions on Neural Networks and Learning Systems},
  title={Domain Adversarial Reinforcement Learning for Partial Domain Adaptation}, 
  year={2020},
  volume={},
  number={},
  pages={1-15},
  doi={10.1109/TNNLS.2020.3028078}}

@inproceedings{cao2019learning,
  title={Learning to transfer examples for partial domain adaptation},
  author={Cao, Zhangjie and You, Kaichao and Long, Mingsheng and Wang, Jianmin and Yang, Qiang},
  booktitle={Proceedings of the IEEE/CVF Conference on Computer Vision and Pattern Recognition},
  pages={2985--2994},
  year={2019}
}

@inproceedings{choudhuri2020partial,
  title={Partial Domain Adaptation Using Selective Representation Learning For Class-Weight Computation},
  author={Choudhuri, Sandipan and Paul, Riti and Sen, Arunabha and Li, Baoxin and Venkat eswara, Hemanth},
  booktitle={2020 54th Asilomar Conference on Signals, Systems, and Computers},
  pages={289--293},
  year={2020},
  organization={IEEE}
}

@inproceedings{he2016deep,
  title={Deep residual learning for image recognition},
  author={He, Kaiming and Zhang, Xiangyu and Ren, Shaoqing and Sun, Jian},
  booktitle={Proceedings of the IEEE conference on computer vision and pattern recognition},
  pages={770--778},
  year={2016}
}

@inproceedings{tzeng2017adversarial,
  title={Adversarial discriminative domain adaptation},
  author={Tzeng, Eric and Hoffman, Judy and Saenko, Kate and Darrell, Trevor},
  booktitle={Proceedings of the IEEE conference on computer vision and pattern recognition},
  pages={7167--7176},
  year={2017}
}

@inproceedings{venkateswara2017deep,
  title={Deep hashing network for unsupervised domain adaptation},
  author={Venkateswara, Hemanth and Eusebio, Jose and Chakraborty, Shayok and Panchanathan, Sethuraman},
  booktitle={Proceedings of the IEEE conference on computer vision and pattern recognition},
  pages={5018--5027},
  year={2017}
}

@article{hoffman2014lsda,
  title={LSDA: Large scale detection through adaptation},
  author={Hoffman, Judy and Guadarrama, Sergio and Tzeng, Eric S and Hu, Ronghang and Donahue, Jeff and Girshick, Ross and Darrell, Trevor and Saenko, Kate},
  journal={Advances in neural information processing systems},
  volume={27},
  year={2014}
}

@article{yosinski2014transferable,
  title={How transferable are features in deep neural networks?},
  author={Yosinski, Jason and Clune, Jeff and Bengio, Yoshua and Lipson, Hod},
  journal={Advances in neural information processing systems},
  volume={27},
  year={2014}
}

@inproceedings{li2019joint,
  title={Joint adversarial domain adaptation},
  author={Li, Shuang and Liu, Chi Harold and Xie, Binhui and Su, Limin and Ding, Zhengming and Huang, Gao},
  booktitle={Proceedings of the 27th ACM International Conference on Multimedia},
  pages={729--737},
  year={2019}
}

@article{li2020deep,
  title={Deep residual correction network for partial domain adaptation},
  author={Li, Shuang and Liu, Chi Harold and Lin, Qiuxia and Wen, Qi and Su, Limin and Huang, Gao and Ding, Zhengming},
  journal={IEEE transactions on pattern analysis and machine intelligence},
  volume={43},
  number={7},
  pages={2329--2344},
  year={2020},
  publisher={IEEE}
}

@article{zhang2018unsupervised,
  title={Unsupervised domain adaptation using robust class-wise matching},
  author={Zhang, Lei and Wang, Peng and Wei, Wei and Lu, Hao and Shen, Chunhua and van den Hengel, Anton and Zhang, Yanning},
  journal={IEEE Transactions on Circuits and Systems for Video Technology},
  volume={29},
  number={5},
  pages={1339--1349},
  year={2018},
  publisher={IEEE}
}

@article{chen2019complement,
  title={Complement objective training},
  author={Chen, Hao-Yun and Wang, Pei-Hsin and Liu, Chun-Hao and Chang, Shih-Chieh and Pan, Jia-Yu and Chen, Yu-Ting and Wei, Wei and Juan, Da-Cheng},
  journal={arXiv preprint arXiv:1903.01182},
  year={2019}
}

@article{choudhuri2018object,
  title={Object localization on natural scenes: A survey},
  author={Choudhuri, Sandipan and Das, Nibaran and Sarkhel, Ritesh and Nasipuri, Mita},
  journal={International Journal of Pattern Recognition and Artificial Intelligence},
  volume={32},
  number={02},
  pages={1855001},
  year={2018},
  publisher={World Scientific}
}

@article{choudhuri2022coupling,
  title={Coupling Adversarial Learning with Selective Voting Strategy for Distribution Alignment in Partial Domain Adaptation},
  author={Choudhuri, Sandipan and Venkateswara, Hemanth and Sen, Arunabha},
  journal={Journal of Computational and Cognitive Engineering},
  volume={1},
  number={4},
  pages={181--186},
  year={2022}
}

@article{liu2021review,
  title={A review of deep-learning-based medical image segmentation methods},
  author={Liu, Xiangbin and Song, Liping and Liu, Shuai and Zhang, Yudong},
  journal={Sustainability},
  volume={13},
  number={3},
  pages={1224},
  year={2021},
  publisher={MDPI}
}

@article{pan2010survey,
  title={A survey on transfer learning},
  author={Pan, Sinno Jialin and Yang, Qiang},
  journal={IEEE Transactions on knowledge and data engineering},
  volume={22},
  number={10},
  pages={1345--1359},
  year={2010},
  publisher={IEEE}
}

@article{ghifary2016scatter,
  title={Scatter component analysis: A unified framework for domain adaptation and domain generalization},
  author={Ghifary, Muhammad and Balduzzi, David and Kleijn, W Bastiaan and Zhang, Mengjie},
  journal={IEEE transactions on pattern analysis and machine intelligence},
  volume={39},
  number={7},
  pages={1414--1430},
  year={2016},
  publisher={IEEE}
}

@inproceedings{long2013transfer,
  title={Transfer feature learning with joint distribution adaptation},
  author={Long, Mingsheng and Wang, Jianmin and Ding, Guiguang and Sun, Jiaguang and Yu, Philip S},
  booktitle={Proceedings of the IEEE international conference on computer vision},
  pages={2200--2207},
  year={2013}
}

@article{tzeng2014deep,
  title={Deep domain confusion: Maximizing for domain invariance},
  author={Tzeng, Eric and Hoffman, Judy and Zhang, Ning and Saenko, Kate and Darrell, Trevor},
  journal={arXiv preprint arXiv:1412.3474},
  year={2014}
}

@inproceedings{choudhuri2023distribution,
  title={Distribution Alignment Using Complement Entropy Objective and Adaptive Consensus-Based Label Refinement For Partial Domain Adaptation},
  author={Choudhuri, Sandipan and Adeniye, Suli and Sen, Arunabha},
  booktitle={Artificial Intelligence and Applications},
  volume={1},
  number={1},
  pages={43--51},
  year={2023}
}

@book{zhou2012ensemble,
  title={Ensemble methods: foundations and algorithms},
  author={Zhou, Zhi-Hua},
  year={2012},
  publisher={CRC press}
}

@article{dong2020survey,
  title={A survey on ensemble learning},
  author={Dong, Xibin and Yu, Zhiwen and Cao, Wenming and Shi, Yifan and Ma, Qianli},
  journal={Frontiers of Computer Science},
  volume={14},
  pages={241--258},
  year={2020},
  publisher={Springer}
}

@article{han2018co,
  title={Co-teaching: Robust training of deep neural networks with extremely noisy labels},
  author={Han, Bo and Yao, Quanming and Yu, Xingrui and Niu, Gang and Xu, Miao and Hu, Weihua and Tsang, Ivor and Sugiyama, Masashi},
  journal={Advances in neural information processing systems},
  volume={31},
  year={2018}
}

@inproceedings{kim2019nlnl,
  title={Nlnl: Negative learning for noisy labels},
  author={Kim, Youngdong and Yim, Junho and Yun, Juseung and Kim, Junmo},
  booktitle={Proceedings of the IEEE/CVF international conference on computer vision},
  pages={101--110},
  year={2019}
}

@inproceedings{zhang2020distilling,
  title={Distilling effective supervision from severe label noise},
  author={Zhang, Zizhao and Zhang, Han and Arik, Sercan O and Lee, Honglak and Pfister, Tomas},
  booktitle={Proceedings of the IEEE/CVF Conference on Computer Vision and Pattern Recognition},
  pages={9294--9303},
  year={2020}
}

@article{saito2020universal,
  title={Universal domain adaptation through self supervision},
  author={Saito, Kuniaki and Kim, Donghyun and Sclaroff, Stan and Saenko, Kate},
  journal={Advances in neural information processing systems},
  volume={33},
  pages={16282--16292},
  year={2020}
}

@inproceedings{yue2021prototypical,
  title={Prototypical cross-domain self-supervised learning for few-shot unsupervised domain adaptation},
  author={Yue, Xiangyu and Zheng, Zangwei and Zhang, Shanghang and Gao, Yang and Darrell, Trevor and Keutzer, Kurt and Vincentelli, Alberto Sangiovanni},
  booktitle={Proceedings of the IEEE/CVF Conference on Computer Vision and Pattern Recognition},
  pages={13834--13844},
  year={2021}
}

@article{morerio2017minimal,
  title={Minimal-entropy correlation alignment for unsupervised deep domain adaptation},
  author={Morerio, Pietro and Cavazza, Jacopo and Murino, Vittorio},
  journal={arXiv preprint arXiv:1711.10288},
  year={2017}
}

@article{peng2017visda,
  title={Visda: The visual domain adaptation challenge},
  author={Peng, Xingchao and Usman, Ben and Kaushik, Neela and Hoffman, Judy and Wang, Dequan and Saenko, Kate},
  journal={arXiv preprint arXiv:1710.06924},
  year={2017}
}
\end{document}